%% file: main.tex
\pdfoutput=1
\documentclass[11pt]{article}

\usepackage[letterpaper, margin=1.1in]{geometry}
\usepackage{times}
\usepackage[T1]{fontenc}
\usepackage[utf8]{inputenc}
\usepackage{microtype}
\usepackage{amsmath,amssymb}
\usepackage{graphicx}
\usepackage{booktabs}
\usepackage{multirow}
\usepackage{array}
\usepackage{url}
\usepackage{xcolor}
\usepackage{caption}
\usepackage{subcaption}
\usepackage{enumitem}
\usepackage{fancyhdr}
\usepackage[numbers,sort&compress]{natbib}
\usepackage{authblk}
\usepackage[hidelinks]{hyperref}
\usepackage{xurl}

\setlist{nosep,leftmargin=*}

\graphicspath{{figures/}}

\usepackage{placeins}
\makeatletter
\AtBeginDocument{\let\@sub@section\subsection
  \renewcommand{\subsection}{\FloatBarrier\@sub@section}}
\makeatother
\fancypagestyle{plain}{\fancyhf{}\fancyfoot[C]{\thepage}}

\newcommand{\ctx}{\textsc{ctxagent}}
\newcommand{\cc}{Claude Code}
\newcommand{\codex}{Codex}
\newcommand{\oc}{OpenCode}

\title{\bfseries CTXAgent\\[2pt]
\large Self-Controlled Context Management of LLM Agents}

\author[1]{Matthias Busch\thanks{\raggedright Corresponding authors: \texttt{matthias.busch@tuhh.de}, \texttt{christian.cyron@hereon.de}}}
\author[2]{Marius Tacke}
\author[1,3,4]{Roland C. Aydin}
\author[5,6]{Christian Feiler}
\author[1,2]{Christian J. Cyron\protect\footnotemark[1]}

\affil[1]{Institute for Artificial Intelligence and Simulation in Mechanics, Hamburg University
of Technology, Ei\ss{}endorfer Stra\ss{}e, 21073 Hamburg, Germany}
\affil[2]{Institute of Material Systems Modeling, Helmholtz-Zentrum Hereon,
Max-Planck-Stra\ss{}e, 21502 Geesthacht, Germany}
\affil[3]{German Research Center for Artificial Intelligence (DFKI),
Stuhlsatzenhausweg, 66123 Saarbr\"ucken, Germany}
\affil[4]{Department of Materials Science and Engineering, Saarland University,
66123 Saarbr\"ucken, Germany}
\affil[5]{Institute of Surface Science, Helmholtz-Zentrum Hereon, Max-Planck-Stra\ss{}e,
21502 Geesthacht, Germany}
\affil[6]{Institute for Interface Physics and Engineering, Hamburg University of Technology,
Am Irrgarten, 21073 Hamburg, Germany}

\date{}

\begin{document}
\maketitle
\thispagestyle{plain}

\begin{abstract}
Agent harnesses keep long sessions inside the context window by compacting the history when
a token threshold is crossed. We study an alternative in which the model manages its own
context: it decides when to compact, what to keep and what to write as the hand-off, given
two tools and one paragraph of guidance. Since earlier work found that open-weight models
cannot manage their own context without scaffolding, we compare seven frontier models under
identical conditions, with and without the context tools, on token cost and benchmark score,
and we compare our harness with three widely used harnesses on two of these models. The benchmark
is assembled from existing ones: coding exercises, multi-step SciCode problems and
SWE-bench Pro and Terminal-Bench~4.0 tasks are interleaved with planted facts, ephemeral
observations and standing instructions, all checked deterministically, in sessions of
35--212 turns that resemble a user working through different tasks in one chat. Most of the
strongest models manage their context: they compact and curate regularly, the live context
of the mixed sessions stays mostly below 100k tokens, and their benchmark
scores mostly do not drop significantly. For the models that manage their context well, the
cost reduction is substantial but varies between models, because some models compact so
often that other costs increase.
With
Claude Opus~5.5, the self-managed harness matches \cc{} on task success at 34\,\% lower cost
than its explicit 200k-window configuration and 49\,\% lower than its default, with no context
window to choose. We release the agent, the benchmark and all result rows.
\end{abstract}
\textperiodcentered{} context management \textperiodcentered{} context compaction
\textperiodcentered{} prompt caching \textperiodcentered{} agent harnesses

\section{Introduction}

The dominant cost of a long agent session is the prefix
re-sent with every request, and the dominant risk is that the prefix stops fitting or stops
being useful, either because the window overflows or because the information that matters is
buried among tokens that no longer matter. Production harnesses therefore compact: \cc{} trims old tool results, offloads
large ones and, close to the window, replaces the history by a structured
summary~\citep{claudecode2026,anthropiccompaction2026}; \codex{} compacts server-side at a
configured limit and keeps the most recent user messages~\citep{codexcli2026,codexconfig2026};
\oc{} compacts when its configured window is reached~\citep{opencode2026}. In their default
configurations the moment of compaction is a token threshold and its content is a fixed
recipe. The model, which is the only party that knows that a task has just finished and that
the files read for it are no longer needed, is not consulted, and the operator has to choose
a window that suits every task at once.

Whether the model can take over this decision has been examined for open-weight models.
\citet{li2026selfcompact} gave such models a compaction tool and found that, unprompted,
they do not recognise when their own context has degraded; they added a rubric probed at
fixed intervals, admission gates and a forced backstop, and noted that frontier models with
stronger metacognition may not need them. This paper tests that conjecture on seven current
models and separates it from the question it raises next: if a model can manage its context,
what does a harness built on that ability change compared with a harness built on a
threshold? Model-decided compaction has been proposed in several forms, trained into coding
agents~\citep{liu2025cat,zhang2026autocompact,gao2026swemem}, scaffolded with a
rubric~\citep{li2026selfcompact}, or elicited by exposing the context as a
file~\citep{shao2026clm}.

In this study, we add a prompt-only variant of model-managed context: two tools and one
paragraph of guidance, without training or scaffolding, in a local agent, \ctx{}. We run seven
current models under this identical instruction, with and without the tools, on a benchmark
that interleaves coding exercises, multi-step SciCode problems and, in a harder family,
SWE-bench Pro and Terminal-Bench~4.0 tasks with planted facts, ephemeral observations and
standing rules, so that information lost to compaction can become a task failure. We further
run the same model through \cc{}, \codex{}, \oc{} and \ctx{} on byte-identical sessions of
35--212 turns behind one accounting proxy, so that dollar cost
with prompt-cache accounting can be compared across harnesses at their own windows. The paper
reports which models manage their context and how, what the management costs and saves on
each provider's price list, which losses it causes, and what the harness changes for a model
that manages its context well.

\section{Related work}
\label{sec:related}

The documentation of \cc{} states that it
clears older tool outputs first and summarises when the window is near, with a manual
\texttt{/compact} command~\citep{claudecode2026,anthropiccompaction2026};
\citet{liu2026diveclaudecode} describe a five-layer pipeline from the source (budget
reduction, snipping, a time- and cache-aware microcompaction, context collapse,
auto-compaction). \codex{} compacts when the context reaches a configured limit capped at
90\,\% of the window, keeps the most recent user messages up to a token budget, and for hosted
models uses a server-side compaction endpoint~\citep{codexcli2026,codexconfig2026,openaicompaction2026};
on our wire it cut at approximately 145k of an explicit 200k window and at 170--190k of its
default. \oc{} compacts when its configured window is reached; an optional pruning pass is off
by default, and a model-callable compress tool exists only as a third-party plugin that is not
part of the harness as shipped~\citep{opencode2026,opencodedcp2026}. Source-code taxonomies
document how widespread the threshold pattern is~\citep{scaffold2026,barbaste2026harnesseng}.
Two developments are relevant here. Anthropic's API offers, besides threshold compaction, a
compaction-on-demand mode in which the application decides when to
compact~\citep{anthropiccompaction2026}, which is the primitive on which a model-decided
policy can be built; and practitioner tools have adopted the hand-off idea: Amp replaces
compaction by a model-written ``handoff''~\citep{amp2025handoff}, and Cline lets the model
propose a \texttt{new\_task} with a carried-over summary~\citep{baumann2025clinenewtask},
without measuring what it costs or saves.

\citet{lindenbauer2025complexity}
show on SWE-bench that replacing older tool observations by placeholders while keeping
reasoning and tool calls matches LLM summarisation's solve rate at half the cost;
\citet{maskingstale2026} map where that stops helping. Cache economics are a recurring theme:
TokenPilot~\citep{tokenpilot2026} argues that layout mutation breaks prefix continuity and
reports 61--87\,\% cost reductions from cache-efficient management;
CliffCompaction~\citep{nguyen2026cliffcompaction} never summarises a summary, drops rather
than rewrites, reports approximately 51\,\% lower cost on Terminal-Bench against \cc{}'s
auto-compaction as a matched-budget baseline, and finds cache reads to be about 78\,\% of the
uncompacted bill; \citet{lumer2026cache} evaluate caching strategies for long-horizon tasks,
and \citet{satish2026beyondtoken} note that fewer tokens do not reliably mean lower cost or
latency. Slipstream~\citep{slipstream2026} and \citet{min2026trace} validate compaction
against trajectory divergence. \ctx{}'s automatic tiers implement the masking and
cache-stability ideas (Section~\ref{sec:system}); this paper is about the policy above them.

\citet{li2026selfcompact} (SelfCompact) let the model
invoke summarisation with a rubric on when to fire and when to suppress; in their agentic
search setting admission gates and a backstop at 30\,\% of the window are added. Evaluated on
open-weight models (Qwen3 and Qwen3.5, GLM-4.7-Flash, MiniMax-M2.5, MiMo-V2-Flash), it
matches fixed-interval summarisation at 30--70\,\% lower cost, and its limitations section
names GPT-5.5, Claude Opus~4.7 and Gemini~3 Pro as models that may detect context degradation
without the rubric, which is the conjecture tested in Section~\ref{sec:models}. Context as a
Tool~\citep{liu2025cat}, AutoCompact~\citep{zhang2026autocompact}, SWE-MeM~\citep{gao2026swemem},
Focus~\citep{verma2026focus}, AgentFold~\citep{ye2025agentfold} and
Context-Folding~\citep{sun2025contextfolding} make compaction a model action for coding or
web agents, mostly through training. Closest to our setting, Context Language
Models~\citep{shao2026clm} elicit self-managed context zero-shot by exposing the context as a
file; current models including GPT-5.6-Sol, Claude Sonnet~4.6 and Opus~5 then surpass
SelfCompact, Context-Folding and \codex{}-style summaries, while a 9B model has limited
context-management ability. VISTA~\citep{xu2026vista} elicits self-managed context by giving
the model a dashboard of block sizes, recency and remaining budget; our agent shows the model
less: the segment ids of shortened results and the review hint, but no budget. We add to this line a prompt-only variant and two measurements
it has not had: the behaviour of seven current models under one byte-identical instruction,
and dollar cost with prompt-cache accounting against pinned production harnesses at their own
windows.

Harness choice is a hidden variable in agent
evaluation~\citep{kapoor2024agentsmatter,zhang2026harnessdisclosure,lewis2026samemodel}.
\citet{vats2026scaffoldeffect} find that the scaffold changes tokens per solved task by up to
40$\times$ while pass rates move by 0--8 points, and \citet{arino2026identicalruns} find more
than 20$\times$ cost gaps between agents on the same model, mostly through the prompt cache,
with run-to-run variance larger than the differences between pairings. Our harness comparison
has the same shape, in that cost moves while success does not, and the variance finding bounds
what three seeds can establish (Section~\ref{sec:limitations}).

Chat-memory benchmarks probe retention without
tools~\citep{wu2024longmemeval,maharana2024locomo,hu2025memoryagentbench};
MemoryArena~\citep{he2026memoryarena} and SWE-EVO~\citep{le2025sweevo} make later tasks depend
on earlier ones, and LoCoBench-Agent~\citep{qiu2025locobenchagent} adds a multi-session
retention metric to a single-turn predecessor~\citep{qiu2025locobench}. Closest to our probes,
\citet{wang2026lostcompaction} find that side constraints survive compaction only about 17\,\%
of the time, and \citet{chen2026governancedecay} that prohibited tool use rises after
compaction; our standing rules are the same kind of probe. Our instrument differs in that
compaction events are detected on the wire and every probe is labelled with whether a cut lay
between fact and probe, the filler is executable coding work from Aider
polyglot~\citep{gauthier2024polyglot} and SciCode~\citep{tian2024scicode,scicodeverified2026},
and the harnesses are production systems at their own windows.
Context-Bench~\citep{contextbench2026} and Terminal-Bench~\citep{terminalbench2025} are the
independent-task references.

\section{Methods}
\subsection{\ctx{}}
\label{sec:system}

\ctx{} is a local agent with a set of standard tools, extended by a set of context-management tools: \texttt{compact\_context(summary, keep\_from, keep\_ids)}
archives everything between the system prompt and \texttt{keep\_from} and replaces it by one
summary segment whose text the model wrote, without a summariser call or additional cost; an
index of the archived tool results is appended so that a later \texttt{recall} knows its
handle. \texttt{curate\_context(drop=[\{id, note\}])} lists the unreviewed tool results and
lets the model replace the ones it has acted on by its own one-line note; everything not
named is kept verbatim. The engine appends a one-line hint to the newest tool result when a
review is due (every 25 tool batches, or after 20k new tokens); it never acts on the hint
itself. As a backup, a standard automatic compaction is implemented, which acts only as a
safety net.

The guidance is one paragraph of the system prompt (Appendix~\ref{app:prompt}): the agent
manages its context itself and judges what is still relevant; every rewrite invalidates the
prompt cache from the cut, so rewriting after every step is to be avoided;
\texttt{compact\_context} is for task boundaries, when a substantial body of work is finished,
and replaces it by a hand-off summary (goal, decisions and why, files changed, results, open
issues, segment ids worth recalling); \texttt{curate\_context} is for the smaller clutter
within a task; facts the user stated and standing instructions must survive both. The
paragraph is task-agnostic and the same byte string for every model.

\subsection{Benchmark base datasets}
\label{sec:bench}

Two existing benchmarks provide the tasks. Aider-polyglot exercises~\citep{gauthier2024polyglot}
in Python (34), JavaScript (48) and Go (36) are self-contained programming exercises scored by
the shipped tests, one per turn; frontier models solve the Python track at 100\,\%, so the
exercises serve as the work between which facts must survive, and each grows the context by
3--7k tokens. SciCode problems~\citep{tian2024scicode} are multi-step scientific coding
problems, one step per turn, whose later steps call functions written in earlier turns, so
that the conversation is the specification of the later steps. We use the dev split (15
problems, 50 steps), whose reference solutions pass our checker. Four dev steps were passed
only twice in 188 attempts across all arms, models and harnesses; their targets depend on choices the prompt
leaves open (Appendix~\ref{app:unsolvable}). They remain in the scripts so that every arm does
identical work; the ceiling is 46 of 50 steps. The eleven-step Ewald-summation problem is the
hardest chain and recurs in the analysis.

The hard sessions draw on three further sources, chosen
because the mixed sessions leave the strongest models near the ceiling.
Terminal-Bench~4.0~\citep{terminalbench2025} contains 66 tasks with a median expert estimate
of about four hours, each with its own container image and verifier. We use the 22 tasks that
run in a single container within the resources of a 16-core host and that the review by Epoch
AI~\citep{epochtb4review2026} does not flag for scoring defects, and draw nine of them, three
per seed, across the software, science, operations, security, hardware and ML categories.
SWE-bench Pro~\citep{deng2025swebenchpro} provides repository tasks in Python, Go and
JavaScript with hidden tests; from its hard subset of 51 instances we take one Python, one Go
and one JavaScript repository per seed, repositories disjoint across seeds, and within a
repository the instance with the median instruction length. SciCode-Verified
\citep{scicodeverifiedrepo2026} is the corrected test split of SciCode, from which we take
problems of three to five steps, three per seed. Each container task is one turn: the agent
works in the task's own image through \texttt{docker exec} and in the task directory, which
is mounted on the host, and the task's own verifier decides pass or fail.

\subsection{Benchmark assembly}

The sessions are assembled from these tasks such that evidence passes
through the context and later turns depend on earlier ones, with the dependency distance as a
controlled variable. This
rules out an instrument of independent questions answered by single shell commands, on which
accuracy is at ceiling for every harness and no compaction can change a score because no
evidence enters the context.

\begin{figure}[tbp]
  \centering
  \includegraphics[width=\textwidth]{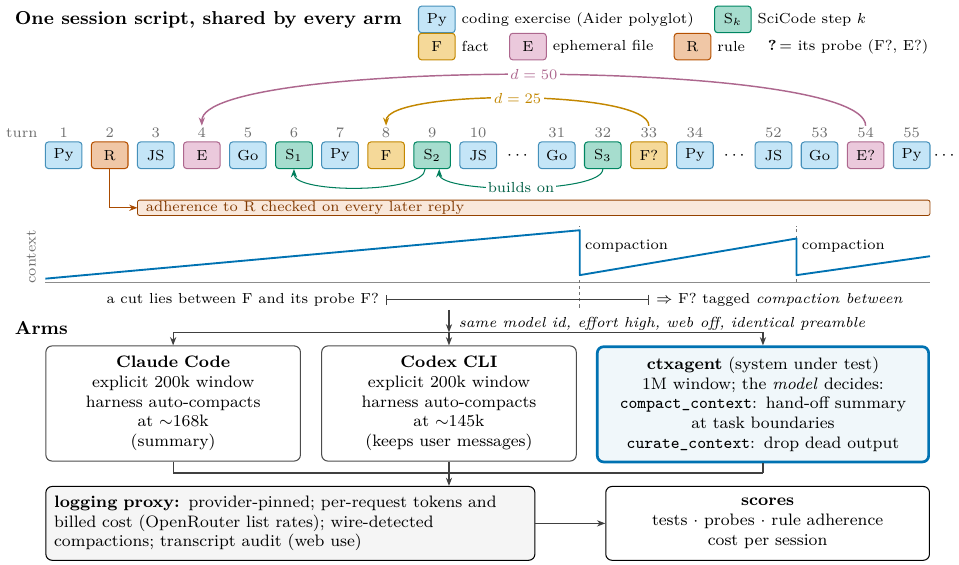}
  \caption{One session script is shared by every arm. Coding exercises and SciCode steps are
  the filler; facts, ephemeral observations and standing rules are planted among them and
  probed at controlled distances; the context trajectory is read from the wire so that every
  probe can be labelled with whether a compaction lay between establishing turn and probe.
  The arms run the same model id through \cc{}, \codex{}, \oc{} and \ctx{} behind one logging
  proxy.}
  \label{fig:design}
\end{figure}

Figure~\ref{fig:design} shows one session script. Three kinds of dependency edge, each with a
deterministic checker, are planted among the filler tasks: stated facts, probed later by a
direct question; ephemeral observations, for which the bench writes a file with a random run
id, asks the agent to print it, and deletes the file before the next turn, so that only the
conversation holds the value; and standing rules (``end every reply with a line \texttt{REF:
<token>}''), checked by regular expression on every later reply. Probe turns instruct the
agent to answer from memory, and the agent's tool calls on probe turns are recorded (zero in
every session). Distances are drawn from $\{3, 10, 25, 50, 90\}$ (up to 200 in the longest
session), every distance at least once per session, and each probe is labelled post hoc with
whether a compaction lay between the establishing turn and the probe, from the compactions
detected on the wire (a prompt-size drop of more than 25\,\% and 10k tokens between
consecutive requests).

Four session families are used. Mixed sessions (\texttt{mixdep:34@s}, three seeds) contain
all 34 Python exercises and 6--7 SciCode problems interleaved in seeded order, 20--23 edges
and two rules, 101--115 turns, 188 scored questions, and 32 fact and 32 ephemeral probes per
arm. Clean chains are five chats of one or two complete SciCode problems and nothing else,
run twice per arm, which give the per-step ceiling without context pressure
(Table~\ref{tab:chains}). The marathon
places everything in one chat: 118 exercises in three languages, all 15 SciCode problems, 21
edges at distances up to 200 and two rules, 212 turns and roughly 1.5M tokens of raw history.
The hard sessions (\texttt{heavy:s}, three seeds with disjoint task sets) interleave three
SWE-bench Pro instances, three Terminal-Bench~4.0 tasks and three SciCode-Verified problems
with seven edges, two rules and three recall probes, which ask from memory for the name of a
function written for an earlier SciCode step: 35--36 turns, of which the six container tasks
are single turns of up to 100 tool calls.
Table~\ref{tab:families} summarises the four families.

\begin{table}[tbp]
\centering\small
\caption{The session families. Edges = planted facts and ephemeral observations with their
probes; every family also plants two standing rules. Arms A and B are \ctx{} with and without
context management at a 1M window.}
\label{tab:families}
\setlength{\tabcolsep}{4pt}
\begin{tabular}{@{}lllp{4.3cm}p{3.6cm}@{}}
\toprule
family & sessions & turns & tasks per session & run with \\
\midrule
Mixed & 3 seeds & 101--115 & 34 exercises, 6--7 SciCode problems, 20--23 edges & \cc{}, \codex{}, \oc{}, \ctx{} (Opus, Sol); seven models, arms A and B \\
Clean chains & 5 chats, 2 reps & 3--11 & 1--2 SciCode problems, no edges & \ctx{} vs.\ \cc{} or \codex{} at 200k (Opus, Sol) \\
Marathon & 1 & 212 & 118 exercises, 15 SciCode problems, 21 edges & \ctx{}, \cc{}, \oc{} (Opus); \codex{}, \oc{}, \ctx{} (Sol) \\
Hard & 3 seeds & 35--36 & 3 SWE-bench Pro, 3 Terminal-Bench~4.0, 3 SciCode-Verified problems, 7 edges, 3 recall probes & five models, arms A and B; \cc{}, \oc{} (Opus) and \codex{}, \oc{} (Sol) at 200k \\
\bottomrule
\end{tabular}
\end{table}

\subsection{Baselines}

\begin{table}[tbp]
\centering\small
\caption{Harnesses and arms. ``native'' is the harness's own window for the model id; 200k
is set explicitly. Web access is off by flag or configuration in every harness, stated in an
identical prompt preamble, and audited in every transcript (zero hits).}
\label{tab:arms}
\begin{tabular}{@{}lllp{5.9cm}@{}}
\toprule
harness & version & windows & compaction \\
\midrule
\cc{} & 2.1.301 & native, 1M, 200k & tool-result trimming; auto-compaction near the window (summary) \\
\codex{} CLI & 0.163.0 & native, 1M, 200k & server-side compaction at window$-13$k; user messages kept \\
\oc{} & 1.19.3 & 200k & auto-compaction when the configured window is reached \\
\ctx{} with compaction & 2a407b9 & 1M & model-decided (\texttt{compact\_context}, \texttt{curate\_context}); age masking as safety net; deep tier at 0.9$\times$1M never reached \\
\ctx{} without compaction & 2a407b9 & 1M & none: engine off, context tools and prompt sections removed (Section~\ref{sec:models} only) \\
\midrule
\multicolumn{4}{@{}p{\textwidth}@{}}{\footnotesize Common to all arms: same model id via
OpenRouter, provider pinned (no fallbacks); reasoning effort high; output reserve
20k; 100 iterations and 1500\,s per turn (3600\,s on the hard sessions); fresh process and configuration per session; order
shuffled by seed; never two trials of the same (model, harness) at once; tools aligned by
class (files, shell, jobs); \ctx{} keeps its context tools additionally.} \\
\bottomrule
\end{tabular}
\end{table}

The self-managed harness is compared with the two vendor CLIs, \cc{} and \codex{}, and the
open-source harness \oc{}, each at the windows listed in Table~\ref{tab:arms}; the model
comparison uses a second \ctx{} configuration with the context management switched off as
the counterfactual. Every request of every harness passes through one logging proxy that
records tokens by class (uncached, cache write, cache read, output) and the provider's billed
cost per request, injects the provider pin, and lets the bench detect compactions and peak
context on the wire independently of what a harness reports. Costs are reported at OpenRouter
list rates on one price table; sessions that ran on OpenAI's flex tier are re-priced per
request from the same tokens. The harness comparison uses Claude Opus~5.5 and GPT-6.1 Sol
(each vendor CLI with its vendor's model, the open-source harness and \ctx{} with both); the
model comparison uses seven models.

\subsection{Metrics and statistics}

Each turn is scored as tests passed, verifier passed (container tasks), probe correct or rule
adhered to. Success is reported over all scored turns and, separately, over the solvable SciCode
steps (46 of 50; 80 step turns per arm over the three mixed sessions) and, on the hard sessions,
over the SciCode-Verified steps and container tasks (51 per arm), because the exercises and
the memory probes are at ceiling for every model and the dependent steps are the part of the
instrument on which models and arms differ. Retention is $P(\text{probe correct}\mid d, \text{compaction between})$. Cost is reported per
session with the token anatomy, compactions per session, peak and mean prompt size, requests,
tool calls and wall time. The unit of randomisation is the session; paired comparisons use the
same (seed, turn) across arms with exact McNemar tests.
Harness errors are excluded and the session is repeated; a cost-cap hit is scored 0 and kept
(none occurred).

\section{Results}

\subsection{Model comparison}
\label{sec:models}

To benchmark how well the models can manage their context window, we ran seven models through
\ctx{} on the three mixed sessions, and five of them on the three hard sessions, in two arms:
arm A, in which the model may compact and curate, and arm B, with the context tools removed
and the two prompt sections dropped, so that the context grows monotonically.

\begin{figure}[tbp]
  \centering
  \includegraphics[width=\textwidth]{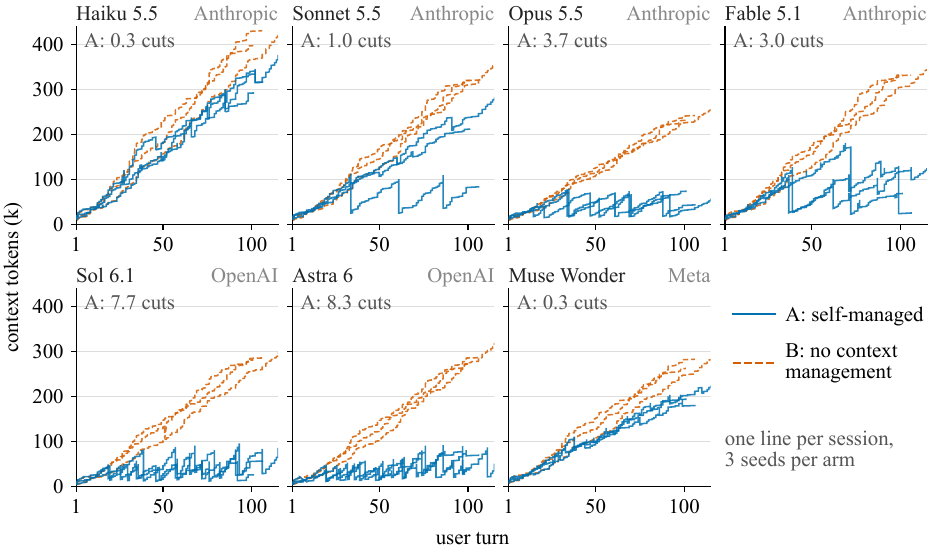}
  \caption{Context size per request against the user turn for seven models under the identical
  instruction to manage their own context (arm A, solid blue) and with no context management
  (arm B, dashed orange), three seeds each. Opus, Sol and Astra cut; Sonnet in one session of
  three; Fable through many small drops; Haiku and Muse Wonder never.}
  \label{fig:p5traj}
\end{figure}

\begin{figure}[tbp]
  \centering
  \includegraphics[width=\textwidth]{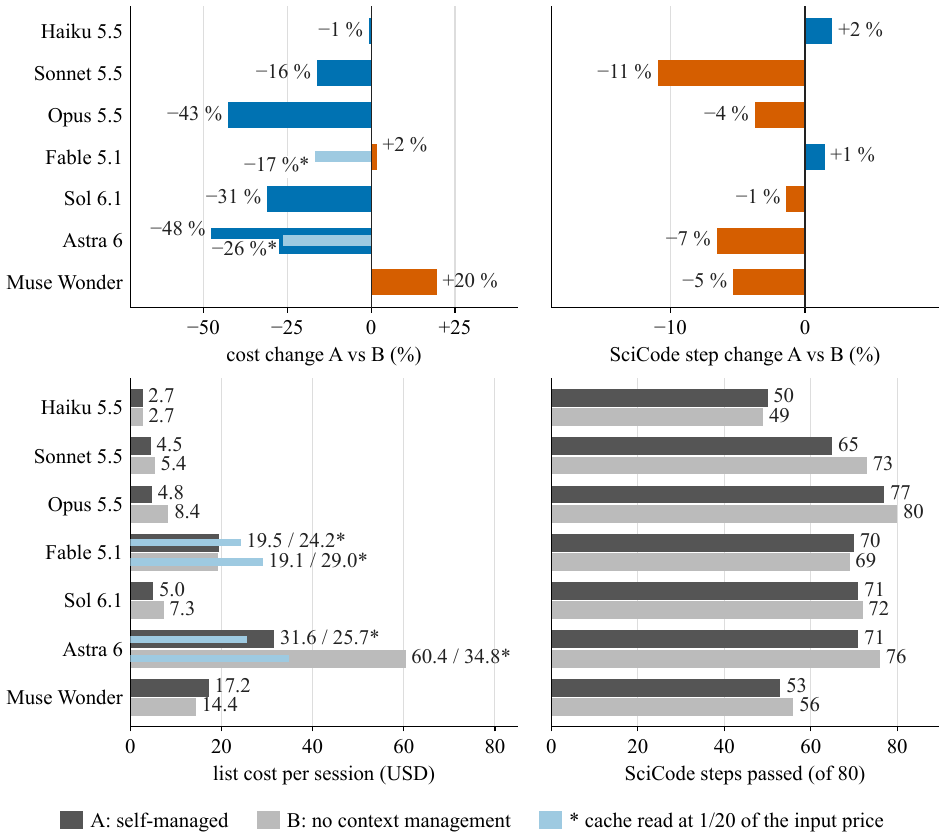}
  \caption{Mixed sessions, pooled over three seeds. Top: change of arm A (self-managed)
  against arm B (no context management) per model; blue marks a change in favour of arm A,
  vermilion one against it. Bottom: the absolute values, arm A (dark) above arm B (light).
  Left: session cost at list rates. Right: solvable SciCode steps passed out of 80;
  exercises and memory probes are at ceiling in both arms. The light-blue
  bars with an asterisk give the cost recomputed with the cache-read price set to a
  twentieth of the input price, the ratio shared by Opus, Sonnet, Sol and Muse Wonder (on the
  list, Fable's cache reads cost a fortieth of its input, Astra's a tenth). The token anatomy
  behind the cost change is Figure~\ref{fig:p5anatomy} in the appendix.}
  \label{fig:p5cost}
\end{figure}

\begin{figure}[tbp]
  \centering
  \includegraphics[width=\textwidth]{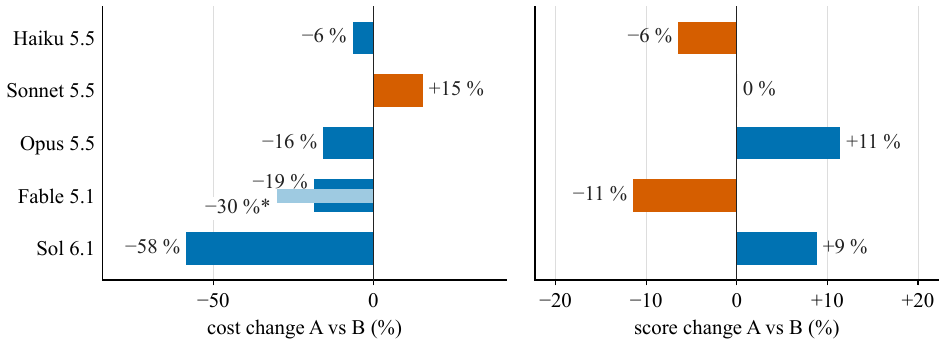}
  \caption{Hard sessions: change of arm A against arm B per model, pooled over three seeds,
  five models. Left: session cost at list rates. Right: SciCode-Verified steps and container
  tasks (SWE-bench Pro, Terminal-Bench~4.0) passed, taken together (51 per arm);
  memory items and probes are at ceiling in both arms. Blue marks a change in favour of
  arm A, vermilion one against it; Fable's light-blue bar with an asterisk is the cost change
  at a cache-read price of a twentieth of the input price, as in Figure~\ref{fig:p5cost}. The
  absolute values are Figure~\ref{fig:p8absolute} and the token anatomy is
  Figure~\ref{fig:p8anatomy}, both in the appendix.}
  \label{fig:p8relative}
\end{figure}

Figure~\ref{fig:p5traj} plots the context size of every request against the user turn, three
sessions per model and arm. The dashed lines are arm B, in which the context grows
monotonically to 230--430k tokens by the end of a session; the solid lines are arm A, with the context tools enabled. Under
the same task strings the models differ in whether, when and how often they act. Compact calls
are the large drops, where the model replaces its history by a hand-off summary it wrote;
curate calls are the smaller and more frequent drops, where it replaces individual tool
results it has acted on by one-line notes.

Opus compacts three to four times per session with a hand-off summary and curates hardly at all. Sol and Astra mix both tools, roughly every ten turns. All three models compact at around 50k--80k tokens, around a third of the occupancy at which \cc{} or \codex{} would cut. Fable acts mostly through
46 small curate drops over three sessions and compacts once or twice per session. Sonnet compacted in
one session of three, with three cuts at 101--112k, and ran the other two to 212k and 280k
without using the tools. Haiku and Muse Wonder never called \texttt{compact\_context} in six
sessions; their 16 and 38 curate calls shrank the context by a few thousand tokens.

Figure~\ref{fig:p5cost} shows the change in cost and in SciCode steps for the mixed sessions, and Figure~\ref{fig:p8relative} shows the same for the hard sessions. Only these results are shown, because the other tasks, the coding exercises and the recall of earlier facts and observations, are at the ceiling in all cases. The standing rules are followed in more than 99\,\% of the replies on the mixed sessions, except by Muse Wonder (83\,\% in arm A, 78\,\% in arm B), and in 80--99\,\% on the hard sessions.
Based on their behaviour and on the change in cost and score, the models fall into three groups. Opus~5.5 and Sol~6.1 reduce the session cost substantially in both settings, by 16--58\,\%, while losing only a few percent of the score on the mixed sessions and improving their score on the hard sessions.

Astra~6 and Fable~5.1, the flagship models, produce mixed results. Like Opus~5.5 and Sol~6.1, Astra and Fable keep their context very small during the mixed sessions, which leads to cost reductions of 26\,\% and 17\,\% at the adjusted cache-read price. The actual costs are strongly affected by the price list: Astra's cached input costs a tenth of fresh input and Fable's a fortieth, whereas Opus, Sonnet, Sol and Muse Wonder charge a twentieth, so the cache reads that arm B re-sends with every request weigh twice as much for Astra and half as much for Fable. At list prices, a similar behaviour therefore yields a reduction of around 50\,\% for Astra and of around 0\,\% for Fable. However, Astra loses 5 of 80 SciCode steps against arm B (71 against 76), while Fable stays nearly constant. On the hard sessions, in turn, Fable loses 11\,\% of its score.

Sonnet~5.5, Haiku~5.5 and Muse Wonder do not manage their context successfully. Sonnet acts in one
session of three: in that session it compacts and saves as Opus does, in the other two it
behaves like arm B, for a mean cost ratio of 0.84.
Muse Wonder and Haiku use only curate calls, which slow the
growth of the context slightly but invalidate parts of the cache with each call. For Haiku
the two effects cancel (ratio 0.99). Muse Wonder costs 20\,\% more; its bill is dominated by
provider-side cache misses that occur with no rewrite at all, to which the curate calls
add.

\subsection{Harness comparison}
\label{sec:harness}

\begin{figure}[tbp]
  \centering
  \begin{subfigure}{\textwidth}
    \centering\includegraphics[width=\textwidth]{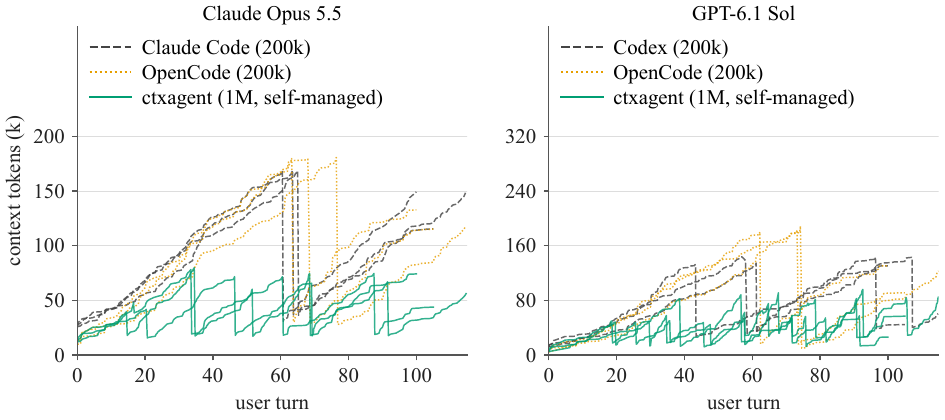}
    \caption{Context size per request on the mixed sessions (three seeds per arm). Left
    Opus~5.5 (\cc{} at 200k dashed, \oc{} at 200k dotted, \ctx{} solid), right Sol~6.1
    (\codex{} at 200k dashed, \oc{} dotted, \ctx{} solid).}
  \end{subfigure}
  \begin{subfigure}{\textwidth}
    \centering\includegraphics[width=\textwidth]{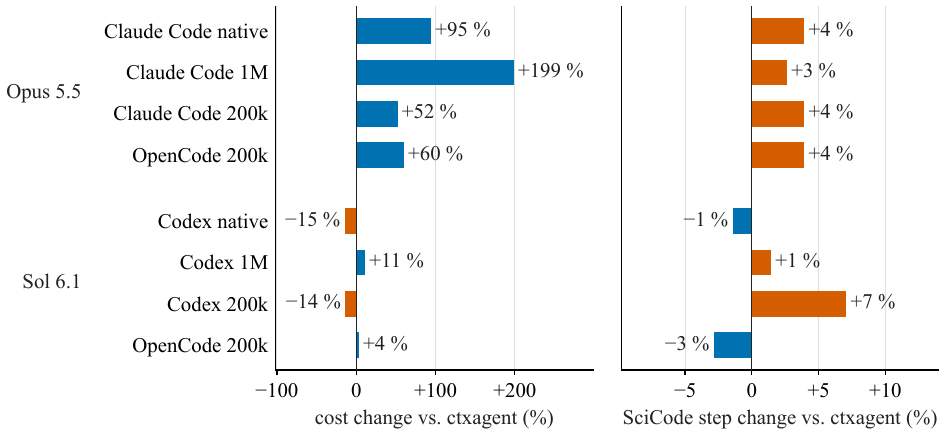}
    \caption{Each harness arm relative to \ctx{} at 1M on the same model, pooled over the three
    seeds: change in list cost per session (left) and in solvable SciCode steps passed out of 80
    (right), including the vendor harnesses at their native and 1M windows and \oc{} at 200k.
    Blue marks a difference in favour of \ctx{} (the arm costs more, or scores lower),
    vermilion one against it.}
  \end{subfigure}
  \caption{Mixed sessions at a fixed model. The threshold harnesses run every session up to
  their cut (\cc{} at approximately 168k, \codex{} at approximately 145k of an explicit 200k
  window); the self-managed arm stays below 100k, with cuts at task boundaries.}
  \label{fig:harness}
\end{figure}

\begin{figure}[tbp]
  \centering
  \includegraphics[width=\textwidth]{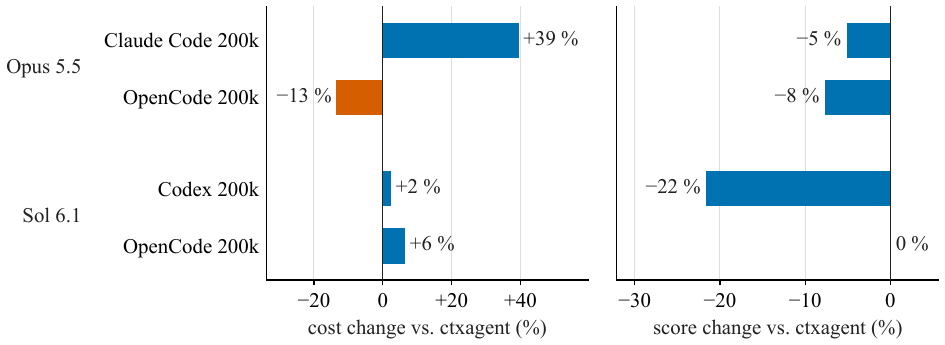}
  \caption{Hard sessions: each harness arm at an explicit 200k window relative to \ctx{} at
  1M (arm A) on the same model, pooled over the three seeds: change in list cost per session
  (left) and in SciCode-Verified steps and container tasks passed, taken together, 51 per arm
  (right). Blue marks a difference in favour of \ctx{}, vermilion one against it.}
  \label{fig:p8harness}
\end{figure}

\begin{figure}[tbp]
  \centering
  \includegraphics[width=\textwidth]{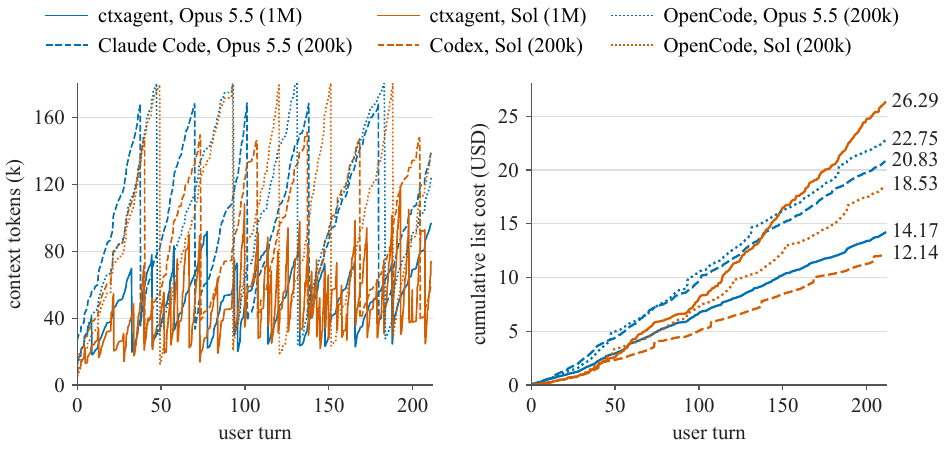}
  \caption{Marathon (212 turns): context size per request (left) and cumulative cost at list
  rates (right); \oc{} at 200k dotted.}
  \label{fig:marathon}
\end{figure}

Figures~\ref{fig:harness} and \ref{fig:marathon} show the context trajectories and the cost
of the same sessions in every harness; Figures~\ref{fig:harnesscost} and \ref{fig:marathonsummary} in the
appendix give the numbers. Across all sessions, the harnesses kept every planted fact and
ephemeral observation except one, which \oc{} lost with Opus on a hard session. Every harness
followed the standing instructions in every reply of the mixed sessions, in more than 99\,\% on
the marathon and in 74--99\,\% on the hard sessions. Task success is also very similar within a
model, the only exception being \codex{} on the hard sessions. The main difference between the
harnesses and models is the cost.

\cc{} did not auto-compact at its default window for Opus~5.5:
its context reached 277k tokens without a cut, and with the \texttt{[1m]} suffix it grew by
7k instead of 3k per turn, reaching 688k and costing 53\,\% more for the same answers. At an
explicit 200k window it compacts once per mixed session at approximately 168k and five times
on the marathon. \ctx{} with the same model never exceeded 80k on the mixed
sessions: it called \texttt{compact\_context} three to four times per session, every time at
an exercise or problem boundary, with hand-off summaries of 1--2k tokens. On the marathon it
compacted eleven times (median 70k tokens) and peaked at 96k. Cost per mixed session for \ctx{} was %
34\,\% below the cheapest tested \cc{} setting
and 49\,\% below the default, at the same number of model
requests (233 vs 228). On the hard sessions and the marathon, \ctx{} likewise cost around 30\,\% less than \cc{}.
\oc{} at 200k with Opus~5.5 behaves similarly to \cc{} at 200k on the mixed sessions and the marathon. On the hard sessions, it is about a third cheaper than \cc{}, roughly at the level of \ctx{}.

The ordering reverses for Sol~6.1. \codex{} at 200k compacts once or twice per mixed
session at approximately 145k; \ctx{} with Sol
compacts about eight times per session at 45--96k tokens, but issues 20\,\% more requests and writes
2.4$\times$ the output tokens. Its hand-off summaries are considerably longer, which eats up the savings: on the mixed sessions it costs 16\,\% more than \codex{} at 200k.
On the marathon, this behaviour is even stronger: \ctx{} produces 5$\times$ the output tokens
of \codex{} and costs \$26.29 against \$12.14. Opus, at twice Sol's per-token price, is the cheaper model in \ctx{} on both the mixed sessions and the marathon.
\oc{} at 200k with Sol~6.1 also compacts at its threshold like \codex{} at 200k, but costs 21\,\% more on the mixed sessions and 53\,\% more on the marathon. On the hard sessions, it scores about 25\,\% higher than \codex{}, the same as \ctx{}.

\section{Discussion}
\label{sec:discussion}

\subsection{Dependence on the model}

\citet{li2026selfcompact} found that open-weight models need scaffolding with a rubric, gates and a backstop to control their compaction.
Section~\ref{sec:models} gives the corresponding picture for a set of current frontier models without that scaffolding and shows that the
context-management capabilities of the models vary considerably. Some models do not use the tools at all or use them inefficiently, while others achieve gains from them.

Optimal context management requires using the right tool at the right time, applying it to the right parts of the context window, and writing precise summaries that keep the important information and leave out the redundant.
To help with the timing, the models receive hints that are not mandatory; the decision when to compact and curate remains theirs. Most models compact much more often than a harness with a 200k window would and still achieve competitive scores, which shows that their summaries are precise enough. However, some models do not compact at all: Haiku~5.5 and Muse Wonder only curate, and not enough to reduce the context significantly. Other models compact so often that the resulting cache invalidation shrinks the cost reduction. A further failure mode is that Sol~6.1 writes very extensive summaries, which increase its output tokens by up to a factor of five. Even so, apart from Muse Wonder on the mixed sessions and Sonnet~5.5 on the hard sessions (15\,\% more at the same score), none of these inefficiencies led to a significant cost increase, nor to a significant drop in benchmark score without a cost reduction of at least similar size. The strongest model in this study, Claude Opus~5.5, managed its context window and reduced its cost significantly while keeping its benchmark scores at the same level on average.

\subsection{\ctx{} against other harnesses}

To analyse how a simple harness with agent-steered context management performs against three widely used harnesses, we
compared \ctx{} in different settings with \cc{}, \codex{} and \oc{}. Again, the results depend on the underlying model,
because different models interact differently with the same system prompt and the same set of tools. This becomes visible in the comparison of \ctx{}
with \oc{} on Opus~5.5 and Sol~6.1: Opus appears to benefit considerably from \ctx{}, with a cost reduction of more than a third on the mixed sessions and the marathon, while with Sol, \ctx{} and \oc{} stay
roughly on par except on the marathon.

In general, \ctx{} with the context tools achieves about the same benchmark results as the baselines. With Claude Opus~5.5, it reduces the cost by 30\,\% on average against the baselines with a 200k window, at about the same benchmark score. Opus~5.5 often reduced its cost to approximately Sol's level, despite twice the per-token price. This shows how much impact intelligent context management can have.

\subsection{Limitations}
\label{sec:limitations}

The results vary strongly, especially where the models decide themselves when and how to compact and curate.
Because of the cost, the number of benchmark runs was very limited, so further, more thorough benchmarking is needed.
The system prompt has not been optimised yet. Instructions that steer the models' compaction and curation behaviour will
likely improve the results further, especially when tailored to the specific model.
Each vendor CLI was run with its vendor's model, so harness and model are confounded in the
cross-vendor cost comparison.
\citet{arino2026identicalruns} report run-to-run cost variance on the same model and harness
that can exceed differences between harnesses: the Opus cost differences here are far outside
the seed spread on the mixed sessions (every \ctx{} seed below every \cc{} seed in each window
condition), whereas
the success differences are not significant. The memory instrument
was at or near ceiling for every arm; it bounds retention from below but does not rank harnesses on
it. Costs are list prices from one provider on one date, and the cache-economics argument
concerns ratios that vendors can change.

\section{Conclusion}

In this study, we examined which language models can manage their own context in
agentic coding sessions and compared a harness built on that ability with three popular harnesses.
The models received two tools to manage their context: a compaction tool that condenses most of the context at once and a curation tool for smaller operations. They received no extensive instructions on how and when to use them, only one short paragraph of broad guidance. We ran
seven current models on identical tasks with and without the tools, and ran the
same model through \cc{}, \codex{}, \oc{} and our agent \ctx{} on identical sessions of
35--212 turns, with planted facts, ephemeral observations, standing rules, dependent SciCode
steps and tasks from Terminal-Bench~4.0 and SWE-bench Pro.

The models differ considerably in their ability to use the compaction and curation tools. Among the seven frontier models, only GPT-6.1 Sol by OpenAI and Claude Opus~5.5 by Anthropic achieved a relatively consistent gain in benchmark score per cost by managing their context themselves. Most other models did not get worse in cost per score, but did not achieve consistent, significant gains either.

The harness comparison showed that our harness \ctx{} is on average at least on par with the other tested harnesses, but the results vary. Relative to the other three harnesses, Opus~5.5 appears to work better with \ctx{} than Sol~6.1 does. The cost reductions of Opus with \ctx{} relative to \cc{} and \oc{} are around 30\,\% or more in most settings, while Sol with \ctx{} mostly stays at the level of the other harnesses.

Whether agent-steered context management benefits task success and cost depends on the model. Under the same instruction, the models behaved differently, ranging from no context management to too frequent management. Among the current frontier models, Opus~5.5 gained most consistently from the context-management tools. GPT-6.1 Sol also improved with the tools enabled, but over long sessions it worked less well and less consistently with our harness than with \codex{} and \oc{}. Further benchmarking is needed, as the results still vary strongly and these benchmarks only partially reflect how a user actually works with an agent. The gains can probably be increased further by optimising the prompt.

\section*{Acknowledgements}
M.B., R.C.A., C.F. and C.J.C. acknowledge funding by the Deutsche Forschungsgemeinschaft (DFG;
project 535656357). We gratefully acknowledge the support of the Helmholtz-Gemeinschaft
Deutscher Forschungszentren (HGF).

\section*{Funding}
This publication was funded by the Deutsche Forschungsgemeinschaft (DFG; project 535656357):
\url{https://gepris.dfg.de/gepris/projekt/535656357}.

\section*{Author contribution: CRediT}
Matthias Busch: Conceptualization, Methodology, Software, Formal analysis,
Investigation, Writing -- Original Draft, Writing -- Review \& Editing, Visualization.
Marius Tacke: Writing -- Review \& Editing.
Roland C. Aydin: Supervision, Writing -- Review \& Editing.
Christian Feiler: Supervision, Writing -- Review \& Editing.
Christian J. Cyron: Supervision, Writing -- Review \& Editing.

\section*{Declaration of competing interests}
The authors declare no competing interests.

\section*{Data and code availability}
The agent, the benchmark code, the harness configurations and the result rows of all reported
sessions are publicly available at \url{https://github.com/MatthiasHBusch/CTXAgent}.

\bibliographystyle{plainnat}
\bibliography{refs}

\appendix

\section{The context-management guidance}
\label{app:prompt}

The paragraph of \ctx{}'s system prompt that constitutes the self-management policy, verbatim
(it follows the generic tool guidelines and a paragraph explaining that shortened tool output
carries a \texttt{segment\_id} and can be restored with \texttt{recall}):

\begin{quote}\small\ttfamily\raggedright
Managing your own context\\
You manage your context yourself: judge what is still relevant to the work ahead and act on
it; the harness only masks old results by age as a safety net. Every rewrite invalidates the
prompt cache from the cut, so do not rewrite after every step. compact\_context is for task
boundaries: when a substantial body of work is finished and the next work does not need its
details, replace it by a handoff summary (goal, decisions and why, files changed, results,
open issues, segment ids worth recalling). curate\_context is for the smaller clutter within
a task: when the harness hints that a review is due, or a step left a lot of dead output, drop
what you have already acted on with a one-line takeaway. Facts the user stated and standing
instructions must survive both.
\end{quote}

Three tool-choice lines are part of the same prompt: ``run\_command for anything that
finishes within about a minute; start\_job for long builds, simulations or servers'';
``Pick the reader by size: read\_file for a file you can read in one call; summarize\_file,
focus\_file or summarize\_folder for large or unfamiliar ones''; and ``Every model round trip
re-sends the whole context: put independent calls in ONE response. Calls run in order, so an
edit and the command that checks it can share one; only a call whose arguments need a result
you have not seen yet must wait.'' The review hint the engine appends reads ``a context review
is due; at the next boundary, \texttt{compact\_context(summary=\ldots)} is the better tool if
the earlier work is finished''. Arm B of Section~\ref{sec:models} drops this paragraph and the
\texttt{recall} paragraph automatically when the engine is off.

The prompt preamble prepended to the first user turn of every session in every harness reads:
``Work only with the files in the working directory and the tools available to you. Do not
access the internet (no web search, no fetching URLs, no package installation, no git clone
from remote). If something seems to require it, say so and continue without it.''

\section{Unsolvable-as-specified SciCode steps}
\label{app:unsolvable}

Steps 44.3 (\texttt{entropies}), 47.4 (\texttt{MC}), 70.8 (\texttt{probabilities\_3nu}) and
78.3 (\texttt{pendulum\_analysis}) of the SciCode dev split were attempted 35, 15, 25 and 5
times across the mixed sessions and clean chains by both models in every harness and never
passed; in the further 108 attempts of the marathon, the \oc{} sessions and the seven-model
comparison, \ctx{}-Opus passed 47.4 once in the marathon and Astra passed 70.8 once. The reference solutions pass the
checker, so the tests are consistent with one implementation; the failures are value
mismatches and, for 78.3, a shape mismatch: the target is a $2\times2$ array while the prompt
asks to ``output the trajectory'' and never states the shape. The common cause is a target
that depends on a choice the step description leaves open: the order of random-number draws
in a Monte-Carlo energy trace, the time-step selection rule and its metric, the convention for
the mixing matrix. The audit log of the SciCode-Verified project~\citep{scicodeverifiedrepo2026}
lists all four steps as defects of the dev split: for 44.3 the reference solution overwrites
the documented input \texttt{d0} with a constant, for 47.4 the tests depend on a shared
random-number state, for 70.8 the third test is numerically ill-conditioned and the reference
solution fails it, and for 78.3 the reference solution selects the time step from the wall
clock so that the output shape is machine-dependent. The corrections released with the paper
itself~\citep{scicodeverified2026} cover only the test split, so no corrected dev steps exist
to re-score. Every arm does the same work on these turns; per-step success is reported on the
46 solvable steps.

\section{Tables}
\label{app:tables}

\begin{figure}[tbp]
  \centering
  \includegraphics[width=\textwidth]{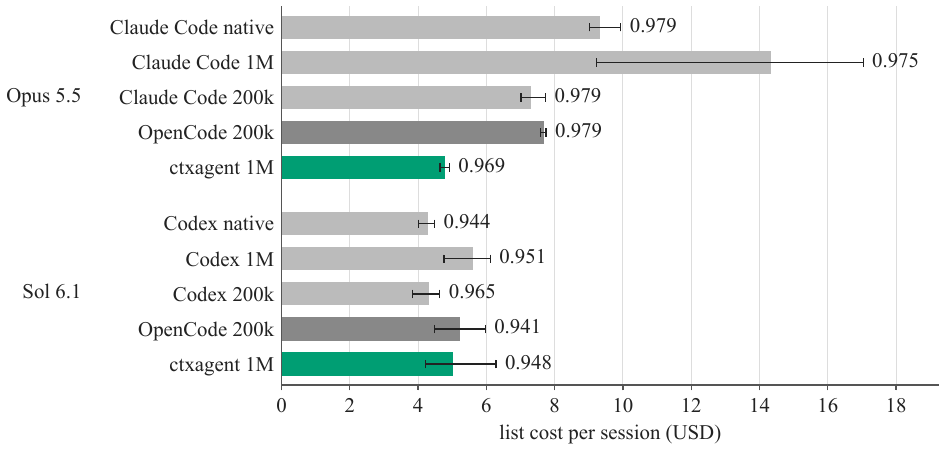}
  \caption{Mixed dependent sessions, three seeds per arm (101--115 turns each): list cost per
  session, mean over seeds with the seed range, for every harness and window; the number at the
  end of each bar is the mean success on the 188 scored questions. Paired McNemar
  \cc{} vs.\ \ctx{} on the same (seed, turn): Opus 3/0, $p = 0.25$; Sol 7/1, $p = 0.07$.
  Memory probes and rules were at 1.0 in every arm.}
  \label{fig:harnesscost}
\end{figure}

\begin{table}[tbp]
\centering\footnotesize
\caption{Five SciCode chats (43 steps per set, 86 over two reps), nothing else in context.
No arm compacted except \ctx{}-Sol. The eight Opus misses per arm are the four
unsolvable-as-specified steps $\times$ two reps.}
\label{tab:chains}
\begin{tabular}{@{}llcccccccr@{}}
\toprule
model & arm & steps /86 & per-chat & \$ (10 chats) & \$/step & peak & cuts & tool calls & wall \\
\midrule
Opus 5.5 & \ctx{} 1M      & 78 & 0.895 & 3.74 & 0.048 & 23--38k & 0 & 128 & 24 \\
Opus 5.5 & \cc{} 200k     & 78 & 0.895 & 4.72 & 0.061 & 38--51k & 0 & 157 & 23 \\
Sol 6.1  & \ctx{} 1M      & 62 & 0.749 & 2.89 & 0.047 & 19--42k & 7 & 276 & 63 \\
Sol 6.1  & \codex{} 200k  & 68 & 0.804 & 2.62 & 0.039 & 22--50k & 0 & 262 & 59 \\
\bottomrule
\end{tabular}
\end{table}

\begin{figure}[tbp]
  \centering
  \includegraphics[width=\textwidth]{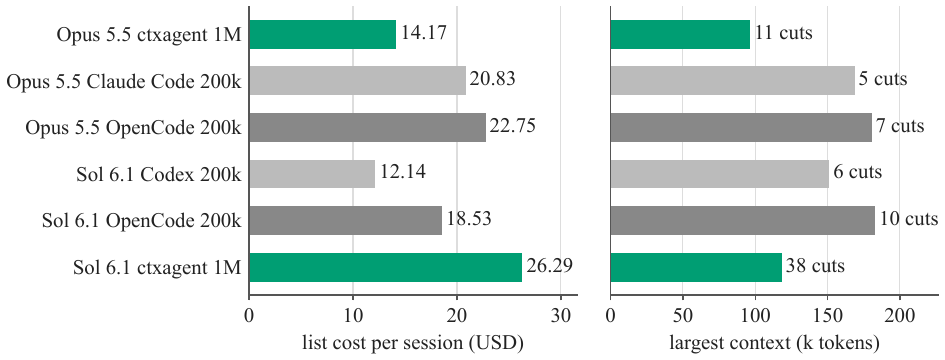}
  \caption{One 212-turn marathon session per arm (118 exercises in three languages, 50
  SciCode steps, 21 edges at distances up to 200, two rules; 168 scored questions). Left: list
  cost of the session. Right: largest context, with the number of cuts. Success was
  0.986 / 0.981 / 0.980 for Opus with \ctx{}, \cc{} and \oc{}, and 0.960 / 0.953 / 0.965 for
  Sol with \ctx{}, \codex{} and \oc{}; every memory item was kept in every arm.}
  \label{fig:marathonsummary}
\end{figure}

\begin{table}[tbp]
\centering\scriptsize
\caption{Seven models, arm A (self-managed) vs arm B (no context management), 3 seeds per
cell, list rates. Success = all scored turns; steps = solvable SciCode steps passed out of
80; memory = facts, ephemerals, probes, rules. Occupancy = prompt size of the last request
before a cut, median [range]; a cut is a drop of the context between consecutive requests as in Figure~\ref{fig:p5traj}. Prices in \$/M tokens: input / cache read / cache write /
output. Context calls over three sessions: cmp = \texttt{compact\_context}, cur =
\texttt{curate\_context}, rec = \texttt{recall}.}
\label{tab:models}
\setlength{\tabcolsep}{2.5pt}
\input{tables/p5_models.tex}
\end{table}

\begin{figure}[tbp]
  \centering
  \includegraphics[width=\textwidth]{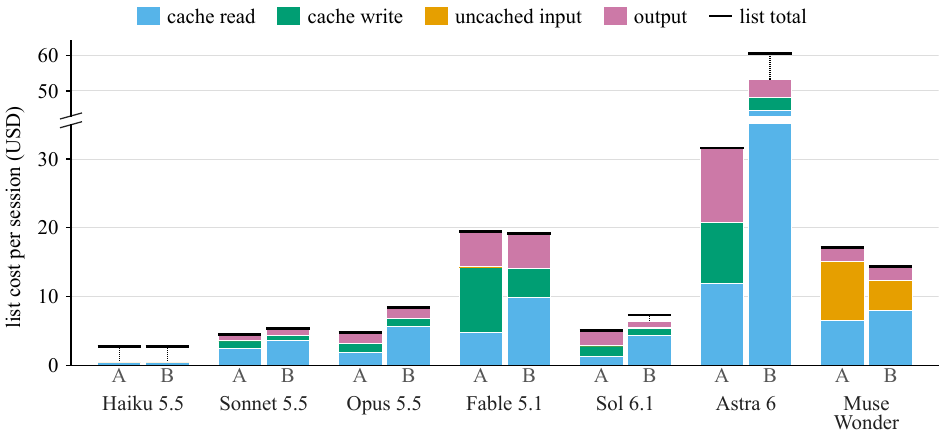}
  \caption{Token anatomy of the session cost on the mixed sessions (mean of three seeds) for
  arms A and B of the seven-model experiment: cache read, cache write, uncached input and
  output at list prices. The black marker gives the list cost including the long-prompt tiers
  (Haiku above 100k context tokens, OpenAI above 272k), which the flat split omits; the dotted
  line spans the surcharge.}
  \label{fig:p5anatomy}
\end{figure}

\begin{figure}[tbp]
  \centering
  \includegraphics[width=\textwidth]{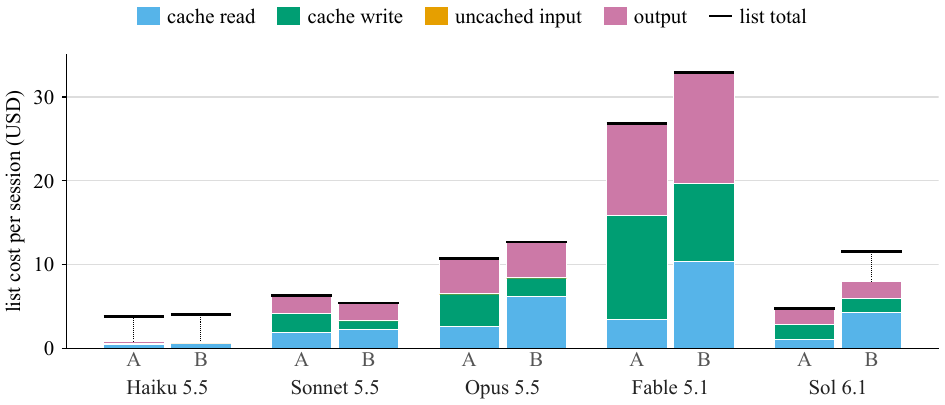}
  \caption{Token anatomy of the session cost on the hard sessions (mean of three seeds), five
  models, arms A and B, as in Figure~\ref{fig:p5anatomy}.}
  \label{fig:p8anatomy}
\end{figure}

\begin{figure}[tbp]
  \centering
  \includegraphics[width=\textwidth]{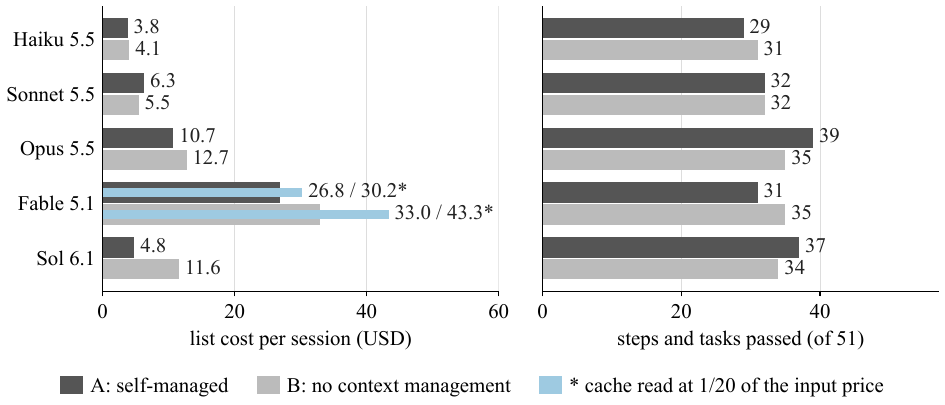}
  \caption{Hard sessions: the absolute values behind Figure~\ref{fig:p8relative}, arm A
  (dark) above arm B (light). Left: list cost per session, mean of three seeds. Right:
  SciCode-Verified steps and container tasks passed, taken together, pooled over three seeds
  (51 per arm). Fable's light-blue bars with an asterisk give the cost at a cache-read price
  of a twentieth of the input price, as in Figure~\ref{fig:p5cost}.}
  \label{fig:p8absolute}
\end{figure}

\begin{table}[tbp]
\centering\scriptsize
\caption{Hard sessions: five models, arm A (self-managed) vs arm B (no context management),
three seeds with disjoint task sets, list rates. Success = all scored turns; SWE = SWE-bench
Pro instances solved of 9; TB = Terminal-Bench~4.0 tasks solved of 9; steps = SciCode-Verified
steps passed of 33; memory = facts, ephemerals, probes, rules; peak = largest prompt over the
three sessions; context calls over three sessions as in Table~\ref{tab:models}. Below: outcome
of every container task, arm A/arm B, 1 = the task's verifier passed.}
\label{tab:hard}
\setlength{\tabcolsep}{2.5pt}
\input{tables/p8_hard.tex}
\vspace{1em}
\input{tables/p8_tasks.tex}
\end{table}

\begin{table}[tbp]
\centering\scriptsize
\caption{Per-session rows of the mixed dependent sessions (seeds 0--2): the baselines at native,
1M and explicit 200k windows, and \ctx{}. Success = all scored turns; memory = probes,
ephemerals and rules; cuts = compactions detected on the wire.}
\input{tables/mixdep_sessions.tex}
\end{table}

Retention by distance and compaction-between was 1.000 in every cell of every arm; the cell
counts for the mixed sessions are, per arm, 44--45 probe-distance observations at $d=3$,
78--83 at $d=10$, 127--133 at $d=25$, 208 at $d=50$ and 162 at $d=90$, of which (\cc{} at
200k / \ctx{}-Opus / \codex{} at 200k / \ctx{}-Sol) 301 / 507 / 348 / 552 had a compaction
between the establishing turn and the probe.

\end{document}

%% file: tables/p5_models.tex
\begin{tabular}{@{}llccccrrrrclr@{}}
\toprule
model (prices) & arm & success & steps & memory & \$/sess. & A/B & peak & prompt & cuts & occ.\ at cut & context calls & wall \\
\midrule
Haiku 5.5 (0.1/0.01/0.125/0.5) & A & 0.873 & 50 & 1.000 & 2.71 & 0.99 & 339k & 174k & 0.3 & 121k & cur 16, rec 2 & 23 \\
 & B & 0.870 & 49 & 1.000 & 2.72 &  & 417k & 203k & 0.0 & -- & -- & 23 \\
Sonnet 5.5 (2/0.10/2.5/10) & A & 0.926 & 65 & 1.000 & 4.50 & 0.84 & 202k & 107k & 1.0 & 110k [101--112k] & cmp 3 & 20 \\
 & B & 0.954 & 73 & 1.000 & 5.37 &  & 333k & 163k & 0.0 & -- & -- & 22 \\
Opus 5.5 (4/0.20/5/20) & A & 0.968 & 77 & 1.000 & 4.80 & 0.57 & 77k & 42k & 3.7 & 72k [40--80k] & cmp 11, cur 2, rec 3 & 21 \\
 & B & 0.979 & 80 & 1.000 & 8.37 &  & 244k & 127k & 0.0 & -- & -- & 24 \\
Fable 5.1 (10/0.25/12.5/50) & A & 0.944 & 70 & 1.000 & 19.46 & 1.02 & 151k & 75k & 3.0 & 112k [55--174k] & cmp 5, cur 46 & 34 \\
 & B & 0.940 & 69 & 1.000 & 19.14 &  & 337k & 169k & 0.0 & -- & -- & 31 \\
Sol 6.1 (2/0.10/2.5/10) & A & 0.947 & 71 & 1.000 & 5.03 & 0.69 & 90k & 39k & 7.7 & 57k [45--96k] & cmp 23, cur 17 & 66 \\
 & B & 0.951 & 72 & 1.000 & 7.30 &  & 289k & 138k & 0.0 & -- & -- & 45 \\
Astra 6 (10/1.00/12.5/50) & A & 0.951 & 71 & 1.000 & 31.61 & 0.52 & 90k & 38k & 8.3 & 58k [22--92k] & cmp 25, cur 10 & 48 \\
 & B & 0.965 & 76 & 1.000 & 60.39 &  & 293k & 139k & 0.0 & -- & -- & 38 \\
Muse Wonder (4/0.20/--/20) & A & 0.884 & 53 & 1.000 & 17.16 & 1.20 & 207k & 110k & 0.3 & 111k & cur 38 & 27 \\
 & B & 0.887 & 56 & 0.985 & 14.35 &  & 277k & 138k & 0.0 & -- & -- & 27 \\
\bottomrule
\end{tabular}

%% file: tables/p8_hard.tex
\begin{tabular}{@{}llcccccrrrrlr@{}}
\toprule
model & arm & success & questions /51 & SWE /9 & TB /9 & steps /33 & memory & \$/sess. & A/B & peak & context calls & wall \\
\midrule
Haiku 5.5 & A & 0.760 & 29 & 6 & 1 & 22 & 1.000 & 3.79 &  & 554k & cmp 1, cur 8, rec 1 & 53 \\
 & B & 0.782 & 31 & 8 & 2 & 21 & 1.000 & 4.05 & 0.94 & 625k & -- & 61 \\
Sonnet 5.5 & A & 0.793 & 32 & 7 & 4 & 21 & 1.000 & 6.29 &  & 308k & cmp 1 & 59 \\
 & B & 0.793 & 32 & 7 & 4 & 21 & 1.000 & 5.45 & 1.15 & 366k & -- & 49 \\
Opus 5.5 & A & 0.869 & 39 & 8 & 6 & 25 & 1.000 & 10.72 &  & 222k & cmp 8, cur 10, rec 1 & 55 \\
 & B & 0.826 & 35 & 7 & 6 & 22 & 1.000 & 12.71 & 0.84 & 426k & -- & 50 \\
Fable 5.1 & A & 0.782 & 31 & 7 & 3 & 21 & 1.000 & 26.81 &  & 215k & cmp 7, cur 10, rec 1 & 68 \\
 & B & 0.825 & 35 & 8 & 6 & 21 & 1.000 & 32.95 & 0.81 & 527k & -- & 79 \\
Sol 6.1 & A & 0.847 & 37 & 9 & 4 & 24 & 1.000 & 4.81 &  & 125k & cmp 16, cur 21 & 56 \\
 & B & 0.814 & 34 & 8 & 5 & 21 & 1.000 & 11.57 & 0.42 & 421k & -- & 98 \\
\bottomrule
\end{tabular}

%% file: tables/p8_tasks.tex
\begin{tabular}{@{}lllccccc@{}}
\toprule
seed & source & task & Haiku 5.5 & Sonnet 5.5 & Opus 5.5 & Fable 5.1 & Sol 6.1 \\
\midrule
0 & SWE & ansible/ansible & 1/1 & 1/1 & 1/1 & 1/1 & 1/1 \\
0 & SWE & element-hq/element-web & 0/1 & 1/1 & 1/1 & 1/1 & 1/1 \\
0 & SWE & flipt & 1/0 & 1/1 & 1/0 & 1/1 & 1/1 \\
0 & TB & html-js-filter & 0/0 & 0/0 & 0/0 & 0/0 & 0/1 \\
0 & TB & production-planning & 0/0 & 0/0 & 0/1 & 0/1 & 0/0 \\
0 & TB & roy-polymorph-cn & 0/0 & 0/0 & 1/1 & 0/1 & 0/0 \\
1 & SWE & gravitational/teleport & 1/1 & 1/1 & 1/1 & 1/1 & 1/0 \\
1 & SWE & internetarchive/openlibrary & 0/1 & 0/0 & 1/1 & 0/1 & 1/1 \\
1 & SWE & protonmail/webclients & 1/1 & 0/0 & 0/0 & 0/0 & 1/1 \\
1 & TB & fin-saccr-rwa & 0/0 & 1/1 & 1/0 & 1/1 & 0/1 \\
1 & TB & photonic-waveguide-routing & 0/0 & 0/0 & 0/0 & 0/0 & 1/1 \\
1 & TB & sound-change-cascade & 0/0 & 1/1 & 1/1 & 1/1 & 1/1 \\
2 & SWE & NodeBB/NodeBB & 1/1 & 1/1 & 1/1 & 1/1 & 1/1 \\
2 & SWE & navidrome/navidrome & 0/1 & 1/1 & 1/1 & 1/1 & 1/1 \\
2 & SWE & qutebrowser/qutebrowser & 1/1 & 1/1 & 1/1 & 1/1 & 1/1 \\
2 & TB & cad-model & 0/0 & 0/0 & 1/1 & 0/0 & 1/0 \\
2 & TB & react-lead-form & 0/1 & 1/1 & 1/1 & 0/1 & 0/0 \\
2 & TB & risk-scorer-replay & 1/1 & 1/1 & 1/1 & 1/1 & 1/1 \\
\bottomrule
\end{tabular}

%% file: tables/mixdep_sessions.tex
\begin{tabular}{@{}lllrrrrrrrr@{}}
\toprule
model & arm & seed & success & memory & list \$ & requests & mean prompt (k) & peak (k) & cuts & wall (min) \\
\midrule
Opus 5.5 & Claude Code (native) & 0 & 0.966 & 1.000 & 9.02 & 220 & 142 & 263 & 0 & 23 \\
 &  & 1 & 0.989 & 1.000 & 9.08 & 223 & 141 & 255 & 0 & 24 \\
 &  & 2 & 0.980 & 1.000 & 9.94 & 235 & 148 & 277 & 0 & 22 \\
Opus 5.5 & Claude Code (1M) & 0 & 0.966 & 1.000 & 17.04 & 214 & 289 & 654 & 0 & 22 \\
 &  & 1 & 0.989 & 1.000 & 9.24 & 221 & 146 & 259 & 0 & 21 \\
 &  & 2 & 0.970 & 1.000 & 16.74 & 236 & 255 & 688 & 0 & 20 \\
Opus 5.5 & Claude Code (200k) & 0 & 0.966 & 1.000 & 7.20 & 219 & 94 & 169 & 1 & 27 \\
 &  & 1 & 0.989 & 1.000 & 7.01 & 229 & 88 & 169 & 1 & 28 \\
 &  & 2 & 0.980 & 1.000 & 7.73 & 237 & 92 & 169 & 1 & 27 \\
Opus 5.5 & ctxagent 1M & 0 & 0.966 & 1.000 & 4.84 & 221 & 44 & 78 & 3 & 20 \\
 &  & 1 & 0.989 & 1.000 & 4.64 & 228 & 41 & 80 & 4 & 21 \\
 &  & 2 & 0.951 & 1.000 & 4.92 & 249 & 39 & 72 & 4 & 22 \\
Sol 6.1 & Codex (native) & 0 & 0.966 & 1.000 & 4.02 & 246 & 85 & 189 & 1 & 41 \\
 &  & 1 & 0.925 & 1.000 & 4.38 & 262 & 86 & 169 & 1 & 49 \\
 &  & 2 & 0.941 & 1.000 & 4.48 & 280 & 86 & 187 & 1 & 50 \\
Sol 6.1 & Codex (1M) & 0 & 0.955 & 1.000 & 6.11 & 291 & 136 & 282 & 0 & 56 \\
 &  & 1 & 0.936 & 1.000 & 4.76 & 242 & 132 & 259 & 0 & 49 \\
 &  & 2 & 0.960 & 1.000 & 5.96 & 263 & 134 & 295 & 0 & 63 \\
Sol 6.1 & Codex (200k) & 0 & 0.955 & 1.000 & 3.84 & 258 & 79 & 144 & 1 & 45 \\
 &  & 1 & 0.979 & 1.000 & 4.51 & 308 & 72 & 143 & 2 & 49 \\
 &  & 2 & 0.960 & 1.000 & 4.64 & 296 & 75 & 143 & 2 & 52 \\
Sol 6.1 & ctxagent 1M & 0 & 0.966 & 1.000 & 4.60 & 329 & 40 & 91 & 7 & 38 \\
 &  & 1 & 0.947 & 1.000 & 4.22 & 336 & 35 & 83 & 8 & 48 \\
 &  & 2 & 0.931 & 1.000 & 6.28 & 367 & 42 & 96 & 8 & 112 \\
\bottomrule
\end{tabular}